\documentclass[sigconf]{acmart}

\usepackage[utf8]{inputenc}
\usepackage{amsmath}
\usepackage{multirow}
\usepackage{algorithm}
\usepackage{algpseudocode}
\usepackage{xspace}
\usepackage{xcolor}

\newcommand{\M}{\mathcal{M}}
\newcommand{\D}{\mathcal{D}}

\newcommand{\conf}{\mathsf{conf}}
\newcommand{\guess}{\mathsf{guess}}
\newcommand{\StateAddress}{$\mathsf{State}$\xspace}
\newcommand{\SchoolLevel}{$\mathsf{SchoolLevel}$\xspace}
\newcommand{\MaritalStatus}{$\mathsf{MaritalStatus}$\xspace}
\newcommand{\Education}{$\mathsf{Education}$\xspace}

\AtBeginDocument{%
  }

\copyrightyear{2023} 
\acmYear{2023} 
\setcopyright{acmlicensed}\acmConference[AISec '23]{Proceedings of the 16th ACM Workshop on Artificial Intelligence and Security}{November 30, 2023}{Copenhagen, Denmark}
\acmBooktitle{Proceedings of the 16th ACM Workshop on Artificial Intelligence and Security (AISec '23), November 30, 2023, Copenhagen, Denmark}
\acmPrice{15.00}
\acmDOI{10.1145/3605764.3623905}
\acmISBN{979-8-4007-0260-0/23/11}

\begin{document}

\title{Information Leakage from Data Updates in Machine Learning Models}

\author{Tian Hui}
\affiliation{%
  \institution{The University of Melbourne}
  \country{Australia}}
\email{huith@student.unimelb.edu.au}

\author{Farhad Farokhi}
\affiliation{%
  \institution{The University of Melbourne}
  \country{Australia}}
\email{ffarokhi@unimelb.edu.au}

\author{Olga Ohrimenko}
\affiliation{%
  \institution{The University of Melbourne}
  \country{Australia}}
\email{oohrimenko@unimelb.edu.au}

\renewcommand{\shortauthors}{Hui et al.}

\begin{abstract}
In this paper we consider the setting where machine learning models are retrained on updated datasets in order to incorporate the most up-to-date information or reflect distribution shifts. We investigate whether one can infer information about these updates in the training data (e.g., changes to attribute values of records). Here, the adversary has access to snapshots of the machine learning model before and after the change in the dataset occurs. Contrary to the existing literature, we assume that an attribute of a single or multiple training data points are changed rather than entire data records are removed or added. We propose attacks based on the difference in the prediction confidence of the original model and the updated model. We evaluate our attack methods on two public datasets along with multi-layer perceptron and logistic regression models. We validate that two snapshots of the model can result in higher information leakage in comparison to having access to only the updated model. Moreover, we observe that data records with rare values are more vulnerable to attacks, which points to the disparate vulnerability of privacy attacks in the update setting. When multiple records with the same original attribute value are updated to the same new value (i.e., repeated changes), the attacker is more likely to correctly guess the updated values since repeated changes leave a larger footprint on the trained model. These observations point to vulnerability of machine learning models to attribute inference attacks in the update setting.
\end{abstract}



\maketitle

\section{Introduction}

Machine learning models are shown to leak private information, such as membership in a training dataset \cite{shokri2017membership}, which can be troubling in sensitive application domains. Machine learning models are however not static. Datasets used for training are constantly changing and the model must remain accurate to reflect new trends in data. Therefore, in practice, machine learning models are updated with the arrival of new data to increase accuracy or incorporate distribution shifts in data. Furthermore, various legal frameworks, such as General Data Protection Regulation (GDPR)~\cite{GDPR}, provide users with the right to be forgotten. For machine learning models, this may imply removing the influence of that data point in the model. This can, for instance, be done with machine unlearning~\cite{Cao2015towardsunlearning}. Therefore, many platforms support automatic model updates when new batches of data arrive or data changes. One can incorporate new data in the model either by updating the previous model parameters
using the new data or by retraining the model from scratch to incorporate the effect of data updates. Access to different snapshots of the model across multiple updates can increase privacy leakage~\cite{zanella2020analyzing, Salem2020UpdatesLeak, Jagielski2023combinemultiple}.

In this work, we study the effect of \textit{updates to existing records} rather than the addition of new records between model updates. To this end, we construct attacks that observe the models' updated behaviour (i.e., changes to the confidence scores) to extract information regarding the changes. We first study an attacker who knows which records have been updated and its aim is to infer the value of the updated attribute.  We then study an attacker whose goal is to identify which records have been updated.
Empirical evaluations on two datasets suggest that having black-box access to both the original model and the updated model results in a larger leakage in comparison to having access only to the updated model.

The main contributions are listed below:

\begin{itemize}
    \item We study the problem of inferring information about \emph{updated} training data in machine learning. Compared to previous work we study the setting where some entries of the training data change as opposed to new data being incorporated into the training dataset. 
    \item We propose attack methods for inferring which records are updated and what the updated values are. Our attacks assume black-box access to the models: the attacker is given labels and corresponding confidence scores.
    \item We use two datasets, namely, Census and Lending Club data, to evaluate the success of the proposed attack methods based on two snapshots of the model and to compare them with attacks relying on a single model.
    \item We study conditions in terms of attribute values and update scenarios that render the models most vulnerable to the proposed attacks.
    \item We discuss two defenses: differential privacy and batch update.
\end{itemize}

\section{Threat Model and attack}
\label{sec:threatmodel}

Let $(t, v, y) $ be a record in a dataset, where $t$ is the target (sensitive) attribute,  $v$ are the remaining known attributes, and $y$ is the label. For example, $t$ could be a marriage status, $v$ may contain information about one's date of birth, race and education,and $y$ be the income level. With time, sensitive attribute value may change from $t$ to a new value $t'$ (e.g., marriage status changed from ``unmarried'' to ``married''). We refer to the dataset and the machine learning model trained on it before/after the update as $\D$/$\D'$ and $\M_{\D}$/$\M_{D'}$, respectively.

We consider two attacker goals:
\emph{updated attribute inference} and \emph{updated record inference}. In the first one, the attacker, given~$v$ and~$y$, is interested in inferring the updated attribute value~$t'$. In the second one, the attacker, given~$t,v,y$ of several records, is trying to infer which of these records has been updated. Note that in the latter, the attacker is not interested in guessing~$t'$ but merely guess which record was updated.

Two attack methods are considered.
Both attacks assume black-box access to a model, that is, the attacker can query each model on inputs of its choice and obtain labels predicted by the model and their corresponding confidence intervals.
In the first one, 
the attacker uses only the updated model $\M(\D')$ to carry out its attack.
In the second one, it uses $\M(\D')$ and the original model $\M(\D)$. 
The goal is to examine the difference in attack success rates between these two scenarios and to determine whether access to an additional snapshot of the model can improve the attacker's success rates (i.e., result in a larger information leakage). We assume that the attacker possesses no data distribution knowledge. Therefore the effect is solely from access to the model alone rather than statistical inference from background data distribution knowledge. We now describe each attack method.

\paragraph{Single model attack} This attack scenario is akin to attribute inference~\cite{Jayaraman2022ai}, where the adversary is trying to infer a value of a target attribute. We assume the adversary knows all other attribute values~$v$ and the label value~$y$. The adversary also has access to the model trained on the updated dataset~$\D'$. The adversary's goal is to infer the updated target attribute value~$t'$. We use the black-box attack confidence-based attribute inference (CAI)~\cite{Jayaraman2022ai}. The procedure entails trying all possible values in the domain of the target attribute, denoted as~$T$ (e.g., all marriage statuses), for the target attribute along with the known attribute values~$v$. For each value in~$T$, it records model's confidence on predicting~$y$. The attacker then uses the value that has the highest model confidence for the true label as its guess for~$t'$. This is summarized in Algorithm~\ref{alg:0}. The attack is successful if attacker's guess, $t_\guess$, equals $t'$.

\paragraph{Two model attack} 
This attack considers the output confidence vectors of the two models: before and after the update. Here, the attacker tries all possible values~$T$ of the attribute on the original model and the updated model. It records confidence vectors returned by each model for the true label $y$ for each value in~$T$. It then computes the  difference between the vectors returned by the two models (updated minus original) and picks the value that corresponds to the largest difference, referred to as $t_\guess$. This is summarized in Algorithm~\ref{alg:1}. The attack is successful if~$t_\guess$ equals~$t'$.

\begin{algorithm}
\caption{\label{alg:0}\emph{Single model attack.} Attribute inference attack using the confidence on the true label of the updated model. Let $(t, v, y) $ be a record, where $t$ is the unknown target attribute, $v$ is the remaining known features and $y$ is the label. Let $t'$ be the updated attribute value. Let $T$ denote the set of all possible values for the target attribute. $\M_{\D}(z)$ returns confidence on label $y$ predicted by $\M$ on the data record $z$. The attack is successful if $t_\guess = t'$.}

\begin{algorithmic}[1]
\Require updated model $\M_{\D'}$, where $\D'$ contains updated record $(t',v,y)$. 
\State Initialize array 
$\conf_{\mathsf{new}}$ of size $|T|$ each
\For {each $t$ in $T$}
    \State $z \gets (t,v)$
    \State $\conf_{\mathsf{new}}[t] \gets \M_{\D'}(z)$
\EndFor
\State $t_\guess$ $\gets \arg \max_{t \in T} \conf_{\mathsf{new}}[t]$
\State \Return $t_\guess$
\end{algorithmic}
\end{algorithm}

\begin{algorithm}
\caption{\label{alg:1}\emph{Two model attack.} Attribute inference attack using the confidence difference on the true label for two models.}

\begin{algorithmic}[1]
\Require model $\M_{\D}$ and updated model $\M_{\D'}$, where~$\D$ and~$\D'$ are the same except~$\D$ contains $(t,v,y)$ and~$\D'$ contains $(t',v,y)$ instead. 
\State Initialize arrays 
$\conf_{\mathsf{old}}$ and $\conf_{\mathsf{new}}$ of size $|T|$ each
\For {each $t$ in $T$}
    \State $z \gets (t,v)$
    \State $\conf_{\mathsf{old}}[t] \gets \M_{\D}(z)$
    \State $\conf_{\mathsf{new}}[t] \gets \M_{\D'}(z)$
\EndFor
\State $t_\guess$ $\gets \arg \max_{t \in T} (\conf_{\mathsf{new}}[t] - \conf_{\mathsf{old}}[t])$
\State \Return $t_\guess$
\end{algorithmic}
\end{algorithm}

\section{Experiments}

\subsection{Datasets}
We evaluate the success of the proposed attacks on the following datasets.

\textbf{Census.} The Census data set is collected by Bargav Jayaraman and Zihao Su\footnote{https://github.com/JerrySu11/CensusData} from the American Community Survey Public Use Microdata Sample files\footnote{https://www.census.gov/programs-surveys/acs/microdata/access.html}. It has 1 676 013 records and 12 attributes. This data set consists of records containing demographic information about individuals, such as sex, income, marital status, and education level, which can be highly sensitive. This data set has been used previously for attribute inference attacks~\cite{Jayaraman2022ai}. We consider two update scenarios: one where \MaritalStatus changes and one where \Education\footnote{The Education attribute was created from the original \SchoolLevel attribute to reduce the granularity.} changes. Both \MaritalStatus and \Education have 5 possible values (which corresponds to the size of $T$ in the previous section). In each scenario, the attacker is trying to infer the update value. The goal of the ML task is to infer whether the annual income is greater than \$90 000 or not.

\textbf{LendingClub.} The LendingClub data set contains records of loan information issued by the peer-to-peer loan company Lending Club\footnote{https://www.kaggle.com/datasets/ethon0426/lending-club-20072020q1}. The dataset contains approved loan records from 2007 to 2020. The loan information recorded includes the borrower’s credit history, loan amount, and interest rate. We follow the feature processing procedure as described in 
this repository \cite{LendingClub}, but remove rows with missing values instead of imputation and use all processed features. The data set has 1 769 947 records and 140 attributes after preprocessing. We consider an update to one sensitive attribute \StateAddress that the attacker is trying to infer. The target attribute \StateAddress has 51 possible values (i.e., the size of $T$).
The goal is to predict whether a loan is \textit{charged off} or \textit{fully paid}.

\subsection{Model Training}
For the Census data, we train a multi-layer perceptron model (MLP) following the same model architecture and hyper-parameters used in previous work by Jayaraman et al.~\cite{Jayaraman2022ai}. The MLP model has two hidden layers. Each layer has 256 neurons and the activation function is ReLu. A training set of 50,000 random records are drawn from all records and separately 25,000 random records are drawn as the test set. The trained model has 89\% train and 85\% test accuracy, which is close to the accuracy reported in~Jayaraman et al.~\cite{Jayaraman2022ai}.

For the LendingClub data, we train a logistic regression model (LR). We sample 10\% of all the records as our training and test records after pre-processing. We use the Sci-kit Learn implementation of logistic regression with its default ``lbfgs'' solver and l2 regularization. The model converges within 200 epochs. It should be noted that due to strict convexity of the logistic regression problem, the algorithm will converge to the global minimum. Both train and test accuracy is around 80\%, which is close to the performance in the aforementioned data processing project~ \cite{LendingClub}. 

The models are retrained from scratch on the updated datasets.

\begin{table*}
\caption{Success rates of two attack methods described in~Section~\ref{sec:threatmodel} --- single-model access and two-model access --- when an attribute of a single record is updated. The attacker aims to infer the attribute value \emph{after an update}. Two-model attack has higher attack success rate than a single-model on all but one update rule.}
  \label{single_attack}
  \begin{tabular}{l|cc|cc}
    \hline
    \multirow{2}{*}{\textbf{Dataset (Attribute)}} & \multicolumn{2}{c}{\textbf{Update Rule}} & \multicolumn{2}{|c}{\textbf{Attack Success Rates}} \\
    \cline{2-5}
     & \textbf{Before} & \textbf{After} & \textbf{Single model} & \textbf{Two model} \\
    \hline
    \hline
    \multirow{5}{*}{Census (\MaritalStatus)} & married & divorced & .12 & .32 \\ 
     & married & separated & .51 & .43 \\
     & married & widowed & .45 & .52 \\
     & unmarried & married & .06 & .38 \\
     & divorced & married & .05 & .32 \\
        \hline
     \multirow{2}{*}{Census (\Education)} & medium & high & .01 & .31 \\
     & high & higher & .20 & .51 \\
        \hline
     \multirow{10}{*}{LendingClub (\StateAddress)} & CA & NY & .00 & .06 \\ 
     & CA & TX & .00 & .04 \\
     & CA & FL & .00 & .06 \\
     & CA & OH & .00 & .27 \\
     & CA & LA & .00 & .55 \\
     & CA & SD & .00 & .65 \\
     & LA & SD & .00 & .66 \\
     & VA & CA & .00 & .01 \\
     & VA & LA & .00 & .56 \\
     & LA & OH & .00 & .27 \\
     & NC & GA & .00 & .18 \\
    \bottomrule
  \end{tabular}
\end{table*}

\subsection{Attack Types}

We perform two types of attack experiments. 

\subsubsection{Updated Attribute Inference}
We now describe the attack scenarios we evaluate in the next section. We first simulate an update scenario where only one record out of all records is updated. The record is updated by changing only one attribute. We also investigate the success of the attacks when multiple records were updated. We perform this experiment on both Census and LendingClub datasets. In Census data, we investigate when the target attribute of \MaritalStatus or \Education changes. While in LendingClub, target attribute of \StateAddress for the borrower changes.

\subsubsection{Update Record Inference}
We further investigate attacker's ability to pick out which records are changed when multiple unknown records are changed in a single model update.  We run this experiment on the Census dataset for two settings: when a few records are updated and when many records are updated (i.e., representing a distribution shift as we update $\approx$$5.5\%$ of ``married" individuals in the training set the ``divorced" value). In this setting, the attacker knows all the original ``married'' individuals and is asked to pick out which of these records is subsequently changed to be ``divorced''.

\begin{table*}
  \caption{
  Success rates of two attack methods described in~Section~\ref{sec:threatmodel} --- single-model access and two-model access --- when \StateAddress of multiple records in LendingClub dataset is updated. We vary an update set from 1 to 100 records where each record is updated according to the same update rule. Attack success rate of the two-model attack method grows with the size of the update set.}
  \label{multiple_attack}
  \begin{tabular}{cc|c|cc}
    \toprule
\multicolumn{2}{c|}{\textbf{Update Rule}} & 
\multirow{2}{*}{\textbf{Update Size}} & \multicolumn{2}{c}{\textbf{Attack Success Rates}} \\
\cline{1-2}
\cline{4-5}
      \textbf{Before} & \textbf{After} &  & \textbf{Single model} & \textbf{Two model} \\
    \hline
    \hline
 \multirow{3}{*}{CA} &\multirow{3}{*}{ NY} & 1 & .00 & .06 \\ 
 &  & 10 & .00 & .29 \\
  &  & 100 & .00 & .48 \\
 \hline
  \multirow{3}{*}{VA} &  \multirow{3}{*}{CA} & 1 & .00 & .01 \\
      & & 10 & .00 & .16 \\
      & & 100 & .00 & .24 \\
       \hline
       \multirow{3}{*}{LA} &  \multirow{3}{*}{TX} & 1 & .00 & .05 \\
       &  & 10 & .00 & .27 \\
       &  & 100 & .00 & .20 \\
       \hline
       \multirow{3}{*}{NC} &  \multirow{3}{*}{GA} & 1 & .00 & .18 \\
       &  & 10 & .00 & .40 \\
       & & 100 & .00 & .41 \\
    \bottomrule
  \end{tabular}
\end{table*}

\section{Results}
In this section, we present the results of different attack experiments. We summarize our findings below:
\begin{itemize}
    \item Two-model attacks outperform single-model attacks in attribute inference by breaking free from single-model attacks' idiosyncratic prediction behaviors (e.g., over-prediction of rare values).
    \item Data records with rare values are more vulnerable to attacks.
    \item When multiple records are updated with the same original and updated values (i.e., repeated changes), the attacker is more likely to correctly guess the updated value given the partial target record.
    \item The simple two-model attack has limited ability to identify confidently which points are being updated (partly due to the difficulty of distinguishing nearby members and non-members). Nevertheless, the difference in success rates still suggests an increased information leakage from two models in comparison to a single model.
\end{itemize}

\subsection{Updated Attribute Inference}
In this section we measure whether, given an updated record, an attacker can infer the value of the updated attribute. We consider the setting where only one record is updated and the setting where multiple records are updated to the same value.

\paragraph{Single record update}
Table~\ref{single_attack} summarizes the attack success rates for single record updates, where the attacker aims to infer the updated value of a target attribute (e.g., how the marital status of a particular record has changed). The experiments report attack success rates averaged over 100 experiments for Census and 300 experiments for LendingClub datasets, where different records were updated in each experiment. The two-model attack almost always has a higher success rate. 

To investigate the rationale of the poor performance of single model attacks and the disparity in attack success rates, we consider the distribution of the attribute inference predictions versus the actual distribution of values. We found that single-model attacks tend to concentrate their predictions on one or two values, and rarely predict other values. For example, the number of records with ``separated'' and ``widowed'' \MaritalStatus values comprise only $4\%$ of the Census data. However, the single-model attack predicts $40\%$ to be ``widowed'' and $37\%$ to be ``separated'' when applied to all training records. Similarly, the single-model attack predicts only two less populous states in the LendingClub data, ``ME'' (Maine) and ``AR'' (Arkansas), resulting in zero success rates in most update settings. On the other hand, although two-model attacks still to some extent share the idiosyncratic prediction preferences with its single-model counterpart, this behavior is  less evident and the predictions have a higher chance of aligning with the actual updated value.

We observe that rare values suffer greater attack success rates. For example, ``CA'' (California), is the most common value in the LendingClub data (comprising $14\%$ of the data). When \StateAddress value ``CA'' is updated to another common value ``NY'' (New York, comprising $8\%$ of the dataset), the success rates by two-model attack is $6\%$, but when changed to a rare value ``SD'' (South Dakota, $0.2\%$), the attack success rates rise to $65\%$. This is inline with prior studies showing that membership inference attacks, for example, have higher success rate on records with rare values~\cite{kulynych2019disparate}.

\paragraph{Multiple records update}
Here we evaluate the same attack, but each update now consists of changes to multiple records. All changed records have the same original and updated attribute values. We vary the number of records being changed in a single update to study its effect on the success of the attack. The number of records changed is still small compared to the population, therefore it does not constitute a distribution shift. Table~\ref{multiple_attack} shows the attack success rates for multiple records update. Experiments with update size 1, 10, 100 are repeated 300, 100 and 10 times, respectively.

We observe that compared to single record update above, when multiple records are updated, the attacker is more likely to correctly guess the updated value given the partial target record.

\subsection{Updated Record Inference}
In this experiment, we test whether, given a subset of records, an attacker can infer which records were updated. Compared to experiments in the previous subsection, we do not require the attacker to guess the attribute value of the updated record and only guess that the update has happened. We consider the setting where only a few records are updated and the setting where we simulate a distribution shift by updating many records.

\paragraph{Few records update setting}

We consider a scenario where the attributes and the label of 1000 training data points (before the update) are known to the attacker. We then update 100 of these records by changing the target attribute to new values (unknown to the attacker). The attacker's goal is to guess which 100 records among 1000 original data records have been updated.  Note that in this setting due to the small number of updated records (i.e., 0.2\% of the training data), the change does not constitute a shift in the distribution.

The attack algorithm proceeds as follows. For a given test point (old record), the attacker obtains confidence scores of the original and updated model and computes the difference between the two. The attacker repeats this calculation for all test points (i.e., all 1000 in the experiment described above) and sorts them by the confidence difference. The intuition is that the larger the confidence difference is, the more likely a point has been changed when training the new model.

\begin{figure}[t]
  \centering
  \includegraphics[width=\linewidth]{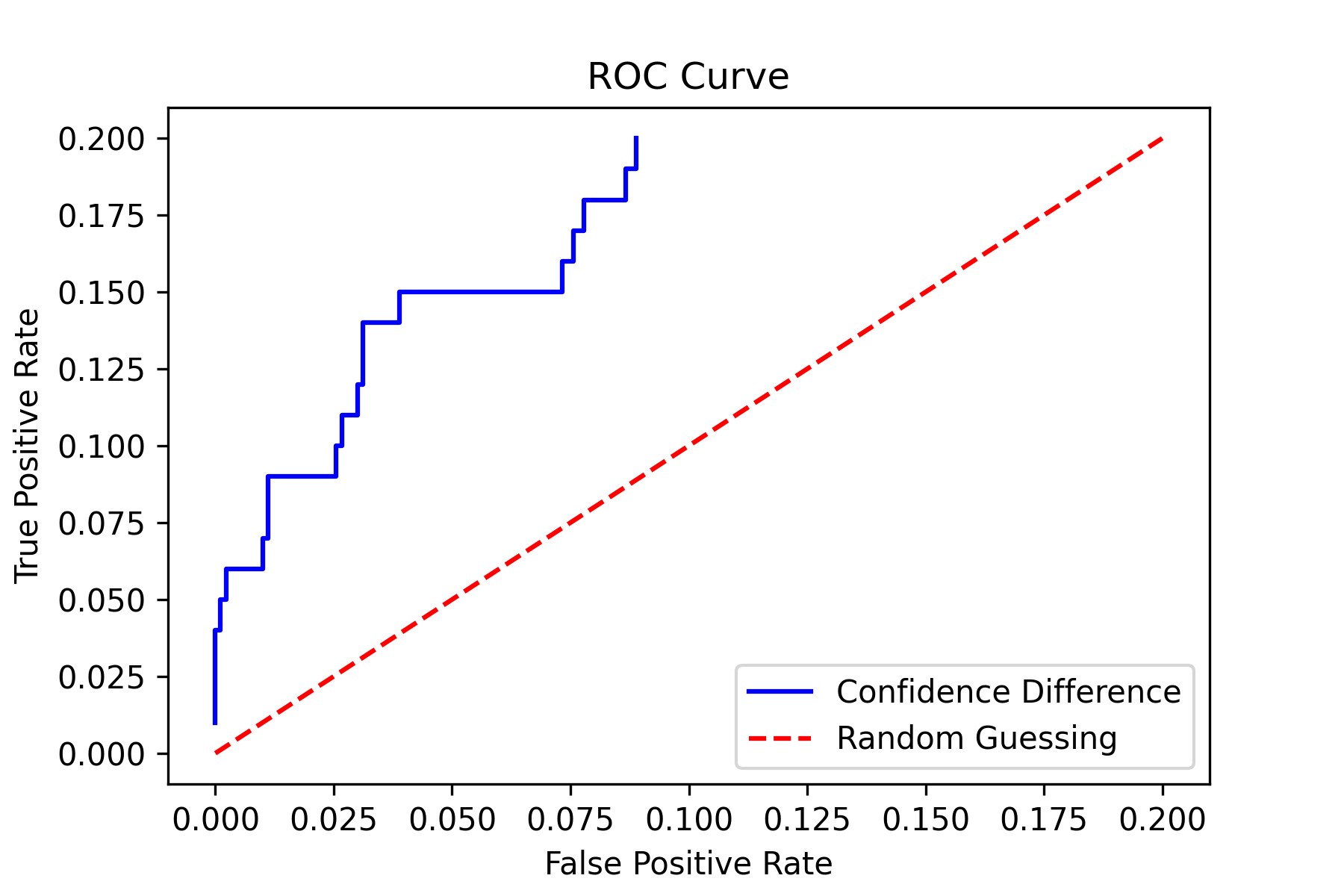}
  \caption{In this experiment, 1000 data records with \MaritalStatus ``married'' are chosen from the Census dataset; \MaritalStatus of 100 of these records is then updated to ``divorced''. An attacker needs to guess which of the 1000 records hav been updated. The attacker can vary the number of their guesses by choosing the top $k$ points sorted by the model confidence difference. By changing $k$, we can vary the false positive rates.}
  \Description{A ROC curve.}
  \label{fig:roc}
\end{figure}

We evaluate the attacker's performance using a receiver operating characteristic (ROC) curve in Figure \ref{fig:roc}. We focus on the low false positive region. This is motivated by the work of Carlini et al.~\cite{carlini2022firstprinciples} that suggests the attacker would ideally aim for high true positive rates at low false positive rates in many realistic settings. It is argued that an attack is considered successful if it \textit{reliably} violates the privacy for even a few users while it is considered unsuccessful if it \textit{unreliably} achieves high aggregate success. Note that the attacker outputs at most 100 guesses. The ROC curve shows that the attack outperforms random guessing. At 100 guesses, the attack picks out~20 true points, which is better than the expected~10 correct points by random guessing. 

We note that this attack is similar to the one on inferring membership between updated models in~\cite{Jagielski2023combinemultiple}. However, our setting is different, since instead of adding or deleting a data record, we update an attribute value of an existing data record and try to infer which record was updated.

\paragraph{Distribution shift update}

We now consider a distribution shift setting where many records are updated. The attacker's goal is again to guess which records have changed while knowing that distribution shift has occurred. In particular, the attacker knows that some records among 27 403 have been updated to ``divorced'' but does not know which ones. In experiments, we change 2 740 ``married'' records to ``divorced'' and refer to this update as a distribution shift since 5.5\% of the training data is updated. 

Table~\ref{pick_out} shows that two-model attacks can pick out more data points that have been updated, but suffer higher false positive rates compared to the results in Figure~\ref{fig:roc}. 
From the perspective of aiming for high true positive rates at low false positive rates, the attacker only showed limited ability to identify which data points have been updated. 
However, the difference in the attacker's behavior may suggest an increased privacy risk with the exposure to two versions of the model. Further work in designing attacks to reduce false positive rates may be interesting.  

\begin{table}[ht]
  \caption{In this experiment, 27 403 data records with \MaritalStatus ``married'' are chosen from the Census dataset; \MaritalStatus of 2 740 of these records is then updated to ``divorced''. An attacker needs to guess which of the 27 403 records have been updated to ``divorced''.}
  \label{pick_out}
  \begin{tabular}{cc|cc}
    \toprule
   \multicolumn{2}{c}{\textbf{TPR}} & \multicolumn{2}{c}{\textbf{FPR}} \\
    \textbf{Single model} & \textbf{Two model} & \textbf{Single model} & \textbf{Two model} \\
    \midrule
     .11 & .35 & .08 & .26 \\ 
    \bottomrule
  \end{tabular}
\end{table}

\subsection{Discussion}
\label{sec:discussion}
We now discuss the influence of randomness in our experiments.
Since logistic regression algorithm converges to global minimum, results  for LendingClub do not depend on randomization. 
However, neural network behavior for Census dataset changes depending on the random seed used for weight initialization and shuffling. The results presented in this section use the same random seed for the original model and the updated model. When the seed is set to be different, the attack success rates drop but still outperform single-model attacks. This is an interesting direction for further research  in terms of investigating various factors of the success of attribute inference attacks using two models and developing \emph{ad hoc} defense mechanisms. As noted in~\cite{hyland2019empirical}, random seeds or randomization in training algorithms in general can have a larger impact on the model than an individual training data point. Therefore, the effect of updating a single data record may be masked by randomization in the training algorithm. To further investigate the attacker's ability to distinguish between the world where a given target data point is indeed updated from one value to another and the world where the target is not updated at all, one can run the same attacks on the two models where different seeds  are used but the training data is not changed.

\section{Defenses}
We discuss two possible defenses: batch update and differential privacy.

\paragraph{Batch Update}
When a large group of records is updated at the same time in a single update, it may be possible for any potential target record to hide among the ``crowd''.
For example, in the data addition setting (as opposed to data change), Zanella-B\'{e}guelin et al.~\cite{zanella2020analyzing} and Jagielski et al.~\cite{Jagielski2023combinemultiple} studied how attack accuracy is affected by model update strategies and sizes of batches with new data. They showed that compared to continued training it is safer to retrain the model from scratch with old and new data; while in the continued training setting the larger batch sizes conceal new data better than small batch sizes.
However, the same cannot be directly applied to the data change setting. Since continued training with updated records cannot reflect data change and erase the footprint of outdated records in previous training, the model needs to be retrained from scratch. However, the idea of large batch updates may still be applicable to our setting based on the following observation. In the experiments we considered multiple data point update for the records that had the same original value and were changed to the same updated value. This was done to measure the contribution of multiple identical changes on the trained model and on the success of the attacks --- which can be seen as the worst-case analysis. This experimental setup is however not entirely reflective of practice as updates can be heterogeneous and changes in opposing directions can potentially cancel the effect of each other on the trained model. Hence, as a defense, one could wait until data updates meet certain heterogeneous conditions and only then update the model.

\paragraph{Differential Privacy}
Differential privacy (DP)~\cite{dwork2006calibrating} provides strong guarantees on the amount of information leakage. It is possible to train both the original model and the updated model with differential privacy to defend against the developed attacks. Jagielski et al.~\cite{Jagielski2023combinemultiple} showed that training with differential privacy at a low privacy budget can offer protection against membership inference attacks on the update set in the data addition setting. However, training with differential privacy can reduce the utility of the model. The purpose of a model update would be to maintain its accuracy and therefore the use of differential privacy is only reasonable if the updated model can still maintain its accuracy in utility-sensitive settings. 

\section{Related Work}
There are existing works on extracting information between updates of models. The attacks can be mainly categorized into the addition and deletion of data points. Various defense methods, including differential privacy, were also proposed.

\paragraph{Membership Inference Attack}
Membership inference attacks aim to determine whether a given data point is in the model training dataset~\cite{shokri2017membership}. Our \emph{updated record inference} attack can be seen as inferring the membership of an individual in the update.

\paragraph{Attribute Inference Attack}
Attribute inference and model inversion attacks~\cite{Fredrikson2015modelinversion} for machine learning models aim to infer the values of an unknown attribute of a record in the training set. Jayaraman et al. \cite{jayaraman2021revisiting} showed that the success of black-box attribute inference attacks on a single model is mostly due to the attacker's knowledge of data distribution, and access to the model has negligible effect. In comparison, we study the  information leakage caused by access to two versions of the model, before and after attribute update. We do not consider the background knowledge of the attacker and study the properties of the model instead.

\paragraph{Attacks in Model Updates}
Zanella-B\'{e}guelin et al. \cite{zanella2020analyzing} showed that training samples used to update a language model can be reconstructed by using differences in prediction scores of the model before and after the update. Jagielski et al.~\cite{Jagielski2023combinemultiple} investigated membership inference attacks between multiple updates and showed that a drastic shift in distribution poses higher risks. Salem et al.~\cite{Salem2020UpdatesLeak} showed how to reconstruct data points in the update set using an attack based on an encoder-decoder architecture. Compared to these previous works, we study the setting where data changes (i.e., attribute values are updated between model updates) as opposed to new data being added.

\paragraph{Machine Unlearning}
Another scenario for model updates is data deletion. In many privacy regulations, such as the European Union's General Data Protection Regulation (GDPR)~\cite{GDPR}, users have the right to request their data be forgotten by the data curator, which could involve machine learning models. Chen et al.  \cite{Chen2021unlearningjeopardizesprivacy} demonstrated that machine unlearning can put the user's data in more danger if an adversary has both versions of models before and after the deletion. Since data deletion can be seen as ``reversed data addition'', the attack methods developed for either of them can be used by swapping the order of the original and updated models. Again, it should be noted that, compared to these  works, we study data changes as opposed to data deletion.

\section{Conclusion}

We considered the case where an adversary has black-box access to machine learning models before and after an update in the training dataset occurs. We showed several settings where an update in an attribute value can result in information leakage. In particular, the change in the model confidence scores can be used to infer which data records and how attribute values in the training data were modified.
This expands the literature on attribute inference in the update setting: we consider a setting when an attribute of a single or multiple training data records is updated rather than entire data records removed or added, which is the current norm in the literature. Using experiments based on two tabular datasets and model families (perceptrons and logistic regression), we demonstrated that access to two snapshots of the model can result in higher information leakage in comparison to having access to only the updated model in various settings. Future work can focus on defending against such attacks by using batch updates and differential privacy.

\section{Acknowledgments}
The work was supported by a seed funding for collaboration between the Department of Electrical and Electronic Engineering (EEE) and  School of Computing and Information Systems (CIS) at the University of Melbourne. We thank Bargav Jayaraman for answering our questions about implementation details and datasets used in~\cite{jayaraman2021revisiting} and~\cite{Jayaraman2022ai}.

\bibliographystyle{ACM-Reference-Format}
\balance
\bibliography{sample-base}

\end{document}